\documentclass[conference]{IEEEtran}

\IEEEoverridecommandlockouts                              
\usepackage{comment}
\usepackage{graphics}
\usepackage{caption}
\usepackage{subcaption}
\usepackage{textcomp}

\usepackage{tikz} 

\usepackage{amssymb}

\usepackage{cite}
\usepackage{wrapfig}

\usepackage{makecell}

\usepackage{algorithm}
\usepackage{algorithmic}

\usepackage{bibentry}
\usepackage{inputenc}
\usepackage{graphicx}
\usepackage{float}
\usepackage{multicol}
\usepackage{multirow}
\usepackage{caption}
\usepackage{amsmath}
\newcommand{\algorithmicinput}{\textbf{\textsl{Input.}}}

\newcommand{\INPUT}{\item[\algorithmicinput]}

\usepackage{threeparttable}
\usepackage{multicol}
\usepackage{breqn}
\renewcommand{\baselinestretch}{0.93}
\usepackage{cleveref}
\usepackage{xspace}

\title{Predictive Probability Path Planning Model For Dynamic Environments}
\author{Sourav Dutta$^{1}$, Tuan Tran$^{1}$, Banafsheh Rekabdar$^{2}$, and Chinwe Ekenna$^{1}$
\thanks{This research is supported in part by NSF awards CRII-IIS-1850319}
\thanks{$^{1}$Sourav Dutta, Tuan Tran and Chinwe Ekenna- Department of Computer Science, University at Albany, SUNY, NY 12206, USA {\tt\small\{sdutta2,ttran3,cekenna\}@albany.edu}.}\thanks{$^{2}$Banafsheh Rekabdar- Department of Computer Science Southern Illinois University, IL 62901, U.S.A. {\tt\small banafsheh.rekabdar@siu.edu}.}%
}

\begin{document}

\maketitle

\begin{abstract}


Path planning in dynamic environments is essential to high-risk applications such as unmanned aerial vehicles, self-driving cars, and autonomous underwater vehicles. 
In this paper, we generate collision-free trajectories for a robot within any given environment with temporal and spatial uncertainties caused due to randomly moving obstacles. We use two Poisson distributions to model the movements of obstacles across the generated trajectory of a robot in both space and time to determine the probability of collision with an obstacle. Measures are taken to avoid an obstacle by intelligently manipulating the speed of the robot at space-time intervals where a larger number of obstacles intersect the trajectory of the robot. Our method potentially reduces the use of computationally expensive collision detection libraries. Based on our experiments, there has been a significant improvement over existing methods in terms of safety, accuracy, execution time and computational cost. Our results show a high level of accuracy between the predicted and actual number of collisions with moving obstacles. 
\end{abstract}

\section{Introduction}
Motion planning is a term used in robotics for finding valid configurations that move a robot from source to destination. An open problem in motion planning is planning in narrow passages and the presence of dynamic obstacles. Methodologies of motion planning find applications in self-driving cars initiatives, cooperative robot swarm scenarios, and efficient mobility of robots in tightly crowded areas while performing important tasks.

 
 \begin{figure}
\centering
\begin{minipage}[b]{0.35\textwidth}
\centering
\includegraphics[width=\textwidth]{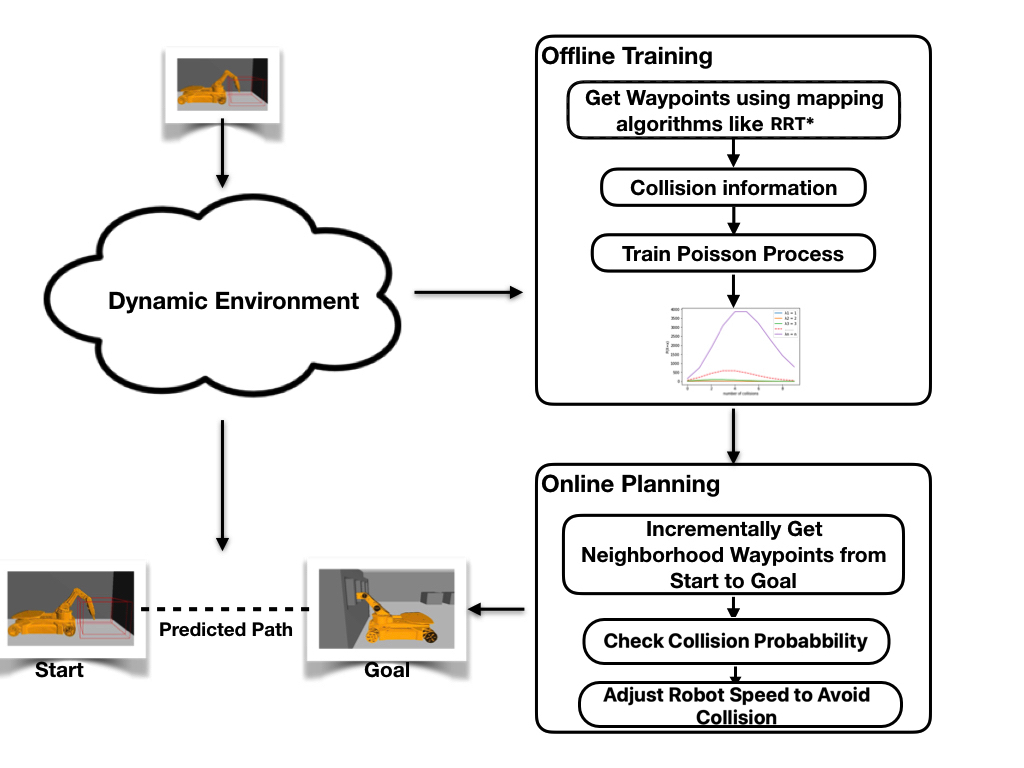}
  \caption{\textbf{Process Description}}
  \label{process_desc}
\end{minipage}
\end{figure}



In robot path planning, spatial uncertainty relates to an area or region of the environment which may or may not be safe to navigate due to complex relationships with moving obstacles leading to a high frequency of collision. Temporal uncertainty defines the time interval during which it becomes uncertain if an identified region is safe to navigate.
In the presence of these uncertainties, it is challenging to generate a successful path for a robot, due to a high probability of collision with moving obstacles.
In robotics, a lot of literature exists to model and alleviate temporal \cite{zhang2017multirobot, lahijanian2016iterative} and spatial \cite{boularias2015grounding, belter2016improving} uncertainties, however, this is still an open research problem. There have been various studies that define spatial and temporal uncertainties over crash data for road safety to prevent motor vehicle accidents \cite{lord2008effects, theofilatos2016predicting, fawcett2017novel}.



Our method, as demonstrated in \Cref{process_desc}, has an offline training and an online planning phase. The online training phase plans a static trajectory for the robot and trains the Poisson processes. The online planning phase estimates the collision probabilities and adjusts the speed of the robot to avoid any collisions.
The \textbf{contributions} of this paper include:

\begin{itemize}
    \item the development of two Poisson random variables that can generate probability mass functions in space and time and predict the probability of collision at any given point of space, or calculate the inter-arrival time of the collision at any given point of time.
    \item a cost-efficient non-homogeneous Poisson process for effective collision detection in dynamic environments.
    \item the creation of a dynamic collision-avoidance framework that can be used with any path planning algorithm.
\end{itemize}

Based on our experiments, we show that our algorithm effectively finds a safe path in a dynamic environment with a high level of accuracy without expensive re-planning.

\section{Related Work}

Collision detection and avoidance in the presence of moving obstacles has received a lot of attention recently. Attempts have been made to solve this problem using a geometric approximation method, an online planning strategy and real-time perception of the environment \cite{goerzen2010survey, yegenoglu1988online, ferguson2015real}. 

\subsection{Geometric Algorithms for fast collision detection}
Researches in creating collision detection libraries most commonly use geometric approximations such as bounding volume hierarchies, inner volume approximation, and workspace certificates. Bounding volume hierarchies, which are used in the collision detection algorithm (Sphere\cite{hubbard1996approximating}, Axis Aligned Bounding Box \cite{tu2009research}, Discrete Orientation Polytopes \cite{klosowski1998efficient}), follow two main steps: decomposing object into regions and bounding those regions with geometric shape primitives. Inner volume approximations pack the geometric shape primitives inside given objects to provide good coverage of the space. The work in \cite{weller2009inner} proposed an irregular sphere packing method or Stolpner et. al. \cite{stolpner2011medial} provided an approach of generating an inner sphere tree by approximation the medial axis. Unlike previous approximation methods, Ghosh \cite{ghoshfast} approximated both obstacle and free workspace with a set of geometric shape primitives and its topology to improve the performance of collision detection. However one of the drawbacks of most of these approximation methods is that they usually use a single type of geometric shape primitives. Thus other approaches rely on the concept of the workspace safety certificate. A workspace safety certificate is regions that are guaranteed to be collision free thus ensure the validity of robot configurations or path segments. Lacevic et. al. \cite{lacevic2016burs} proposed an approach to approximate C-free certificates for efficient path planning. Additionally, Yang et. al. \cite{yang2004adapting} utilized a set of a sphere in the free space to improve the sampling distribution. Other widely used approaches to collision detection are Adaptive Monte Carlo localization on the motion, location and sensors of each agent involved in the collision \cite{broadhurst2005monte,siagian2007biologically,cobano2011path}, this method proved to be computationally expensive. In \cite{hennes2012multi}, Hennes et al. presented a collision-avoidance approach in the presence of moving obstacles, called collision avoidance with localization uncertainty to alleviate the use of Monte-Carlo localization. They used local communication to share the robots' state information to ensure smooth collision-free motion. In the following sections, we discuss the different approaches taken in existing texts to handle moving obstacles to avoid collisions.
\xspace
\subsection{Dynamic collision avoidance using Temporal Logic specification}
In \cite{maly2013iterative}, Maly et al. presented a framework for iterative temporal motion planning in dynamic environments for a hybrid system with complex and nonlinear dynamics given a temporal logic specification in a partially unknown environment. They also defined a scheme to measure the closeness of satisfaction of a temporal logic specification to plan the desired trajectory.
Additionally, Agha-Mohammadi et al. \cite{agha2018slap} proposed a rollout-policy-based algorithm for online replanning in belief space to enable Simultaneous localization and planning (SLAP). The method is able to handle the presence of changes in the environment and large deviations in the robot’s pose by using lazy evaluation of the generated feedback tree and replan accordingly.
Hoxha el at. \cite{hoxha2016planning} proposed a framework for online planning for robotic systems in dynamic environments by utilizing  Metric Temporal Logic specifications to inspect and modify available plans to avoid obstacles and satisfy specifications in a dynamic environment.

\subsection{Poisson Process for collision event counting}
The Poisson process model has been widely used in motion planning fields for collision detection. A closely related work to this paper \cite{hess2013poisson} used a Poisson-driven dirt distribution map for identifying static objects for a cleaning robot.
The environment was tessellated into occupancy grids and each of the cells was observed to learn the rate at which the floor gets dirty. 
A homogeneous Poisson process was constructed for each cell of the grid with the polluting events as a random variable. Given a cleaning operation had taken place at time $s$, the dirt level at time $t > s$ was predicted by the expected value of polluting events. 
In this paper, a similar approach has been used to identify the temporal distribution of moving obstacles, instead of static polluting agents. Therefore our approach requires an intelligent inclusion of spatial random and temporal distribution.
Furthermore, research in \cite{mahmud2016poisson} used the Poisson process, specifically the Log-Gaussian Cox Process, for modeling the inter-activity time to predict the starting time of the next unobserved activity in a video. Using the Poisson process to model the obstacles in the environments, Karaman et al. \cite{karaman2012high} derived lower and upper bounds on the critical speed in which a vehicle could safely traverse those environments. The work in \cite{molina2019go} proposed an exploration approach for mobile robots that builds and refines a spatio-temporal model of pedestrian motion by using the uncertainty of a Poisson process built from past observations to infer the locations and intervals to explore at future times. Moreover, Jovan et. al. \cite{jovan2016poisson} used periodic Poisson processes to count the number of activities performed by a human in a time of different rooms of a building to maximize human-robot interactions. To address the problem of mapping human flows with mobile robots in an indoor environment, Jumel et. al. \cite{jumel2017mapping} used a spatial Poisson process to estimate the human presence probability in each region of that environment.

\subsection{Non-homogeneous Poisson Process}
In \cite{lin2017sampling}, Lin et al. presented a sampling-based motion planning strategy for UAV collision avoidance. They used a closed-loop RRT-based model(Rapidly exploring Random Tree sampling method) with a simplified node connection strategy followed by using intermediate waypoints to create candidate trajectories or by using reachable sets for collision prediction. These intermediate waypoints are defined as milestones or stopping points along the path of travel for the robot where the course of trajectory is subject to change.

Moreover, to address planning problems for human-robot interaction in social environments, Tipaldi el at. \cite{tipaldi2011planning}  learned a non-homogenous spatial Poisson process whose rate function encodes the occurrence probability of human activities in space and time. 
Although this method can be considered suitable for predicting collision with moving obstacles, it contains no information about cells or rarely used and unobserved places in an environment.


While most of the literature discussed above depend on dynamic replanning even after some offline obstacle prediction, in this paper, we use Poisson distribution map to keep track of the number of collisions occurring over distance and time and predict future collision ahead of time and avoid it without any expensive dynamic replanning at the time of traversal when the robot is traveling at a certain speed.

\section{Methodology}

A Poisson distribution is a discrete probability distribution that expresses the probability of a given number of events occurring in a fixed interval of time or space if these events occur with a known constant rate (known as a parameter of the distribution) and their occurrences are independent of the time since the last event. A discrete random variable $X$ is said to be a Poisson random variable with parameter $\lambda$, represented as $X \sim Poisson(\lambda)$, if its range is \{0,1,2,3,...\}, and its probability mass function is given by:

\begin{equation}
    P_X(k) =
\left\{
	\begin{array}{ll}
		e^{-\lambda}\lambda^{k}/k!  & \mbox{for }  k \in \{0,1,2,3,....\}\\
		0 & \mbox{otherwise} 
	\end{array}
\right.
\label{eq1}
\end{equation}

The rate of collision with static obstacles is independent of time but dependent on space. Therefore, they display spatial properties. In this paper, we show that the rate of collision for dynamic obstacles can be dependent on both time and space due to its inherent properties, i.e, constantly changing locations at each timestamp. We implement two Poisson random variables to calculate the rate of collisions over unit distance and unit time. One of the novelties of this algorithm is that both the Poisson random variables are estimated during an offline training phase for an environment with dynamic obstacles. The trajectory generation is also offline, and no replanning strategy has been applied.
\xspace
\subsection{Modeling the environment}
We model the environment comprising of moving obstacles in an infinite Euclidean space with a robot in a finite space boundary. The robot plans its trajectory in a configuration space (C-Space) that comprises of the feasible configuration of the robot defined over vector space. We have modeled the obstacles in relation to the coordinates and no information about the velocity, orientation or trajectory is known beforehand. The obstacles can arrive at random anywhere inside or outside the environment.

\begin{figure}
\centering
\begin{minipage}[b]{0.3\textwidth}
\centering
\includegraphics[width=\textwidth]{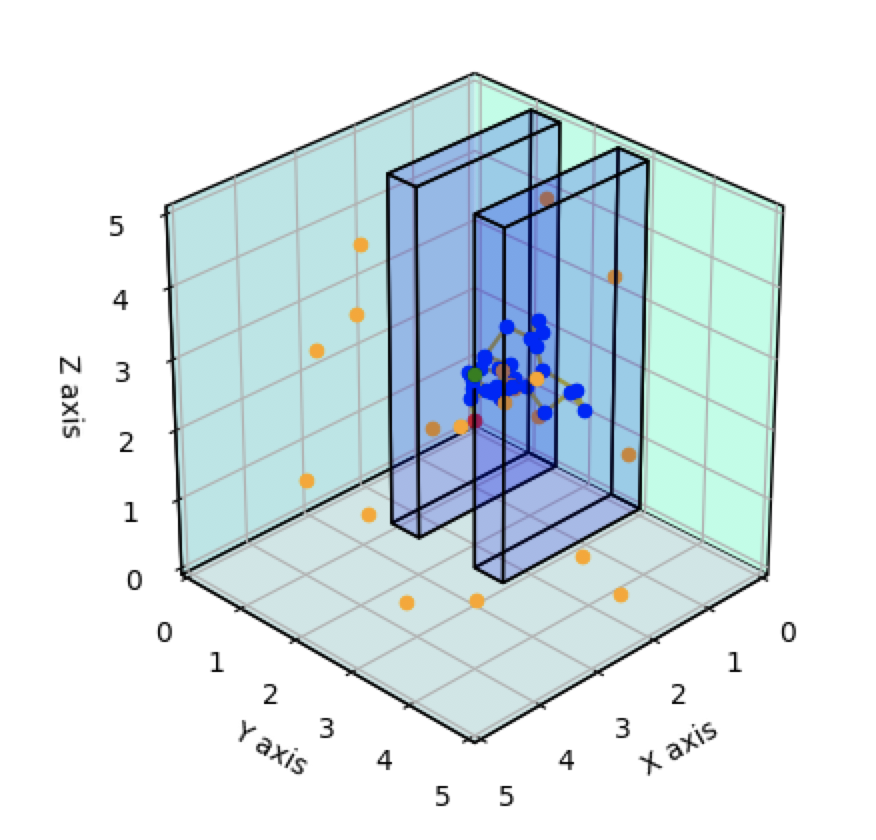}
  \caption{\textbf{Environment with 2 static and 20 random moving obstacles }(marked as floating dots without any edges)}
\label{five_obstacles}
\end{minipage}
\end{figure}

During the training phase, collision counts are determined by the number of times the obstacles appear on the trajectory of the host robot. However, the robot knows about static obstacles in the environment, which it initially plans with. 
\Cref{five_obstacles} shows an environment with 2 static and 20 randomly placed moving obstacles.

\subsection{Spatio-temporal Poisson Process}
A problem with fitting a Poisson Process over a trajectory generated by a path planning algorithm is that the traversal time is independent of the path generation. Therefore, it becomes impossible to observe the number of collisions at the time of path generation. We can, however, generate Poisson parameters over space and over time separately. Our Poisson random variable over space has a parameter estimated over a fixed interval of space is 1-meter distance. A Poisson random variable over time has a parameter estimated over a fixed interval of time of 1 second.

Individually they don't produce a safe path to travel unless we choose to re-plan the path, or know exactly when or where the last collision event occurred, and then calculate the arrival time for the next collision event. That involves a dynamically trained process, which is expensive and may compromise the robot's safety by making it vulnerable to collisions during the training process. However, in order to train a Poisson process offline, we can observe the number of times an obstacle appears on the trajectory of the robot when it is yet to start its traversal of the planned trajectory.

It is not necessary that the dynamic obstacles will be uniformly distributed within the environment. Therefore, the need for a spatial Poisson random variable can be justified in order to focus our observations on specific parts of the trajectories where the probability of collision is high. It can be seen intuitively that in a real-world scenario, dynamic obstacles can only be observed at points of intersection, merger or crossing of multiple roads.

In our method, we have therefore combined two Poisson random variables, one over space and one over time. The spatial variable represents the number of collisions over unit distance and is observed only over the planned trajectory, irrespective of the rest of the environment. 

\begin{equation}
\label{eqspatialparam}
    \lambda=(1/D*n)*\Sigma_{i=1}^{n}C_{i}
\end{equation}

We make observations (like snapshots) of the trajectory at different timestamps and count the total number of collisions (C) over total distance (D) from start to goal. The parameter for the spatial random variable is calculated as in \Cref{eqspatialparam}, where $n$ is the number of observation timestamps: 

The temporal Poisson random variable is used to determine the time-period between each consecutive collisions on the robot's trajectory. We fit a Poisson process on the collision counts over each edge connecting two waypoints and determine the inter-arrival time of the collision events. 
When the robot arrives at a waypoint, it has to determine the time it takes to travel the edge connecting the next waypoint at its current speed. If the time to travel that edge is lesser than the arrival time of the next collision event, then the robot is allowed to pass, else, its speed is decreased and the same probability calculations are repeated. The probability calculations for the temporal random variable has been discussed in \cref{temporal_estimation}.

\subsection{Estimation of Poisson Parameters}
The Poisson parameters for either of the two Poisson random variables were calculated using maximum likelihood estimation (MLE). From the probability mass function in \Cref{eq1}, we get the likelihood function as \Cref{eq3}, where n is the number of desired observations, and $x_i$ is the number of collisions at that timestamp.

\begin{equation}
\label{eq3}
    L(\lambda;x_1,x_2,...,x_n) = \prod^{n}_{i=1}(e^{-\lambda}*\lambda^{x_i})/{x_i}!
\end{equation}

Taking log on both sides we have the log-likelihood function:

\begin{equation}
\label{eq4}
    l(\lambda;x_1,x_2,...,x_n) = \sum^{n}_{i=1}({-\lambda}+{x_i}*log(\lambda)+log(x_i!))
\end{equation}

Taking derivative with respect to $\lambda$:

\begin{equation}
\label{eq5}
    l'(\lambda;x_1,x_2,...,x_n) = -n+(1/\lambda)*\sum^{n}_{i=1}{x_i}
\end{equation}

Setting Equation \ref{eq5} to zero, we have:

\begin{equation}
\label{eq6}
\lambda = (1/n)*\sum^{n}_{i=1}{x_i}
\end{equation}

Equation \ref{eq6} gives the MLE of a Poisson parameter, which is equal to the average of all observations of collision counts (also known as the sample mean).

\subsection{Estimation of the inter-arrival time}
\label{temporal_estimation}
The inter-arrival time has an exponential distribution. The probability density function of the distribution is given as follows:

\begin{equation}
\label{eq7}
    P_X(t) = \lambda*e^{-\lambda*t}
\end{equation}

Therefore the cumulative density function is given as:

\begin{equation}
\label{eq8}
    F_X(t) = \int_{0}^{t} \lambda*e^{-\lambda*t} dx
\end{equation}

Therefore, the probability of the collision event occurring before the robot reaches the next waypoint is determined using (\cref{eq9})

\begin{equation}
\label{eq9}
    P(T \leq t)=F_X(t)
\end{equation}

and subsequently by (\cref{eq2}), where T is the arrival time of the next collision event, t is the time it takes for the robot to travel through the edge, and $\lambda$ is the parameter of the Poisson distribution.

\begin{equation}
\label{eq2}
    P(T \leq t) = 1-e^{-\lambda*t}
\end{equation}

The more reduction in speed that occurs, the greater will be the traversal time $t$ for an edge since our distance between two waypoints is fixed. As can be seen from Equation \ref{eq2}, the probability $P(T\leq t)$ reduces with a decreased speed, hence a collision can be avoided.


\begin{algorithm}
\caption{Offline Training}
\begin{algorithmic}
    \INPUT{env is a $\mathbb{R}^{3+d}$ environment where $d$ is the degrees of freedom of the robot.}
    \INPUT{Start and Goal are the start and goal points of the robot in env.}
    \INPUT{$n$ is the allocated number of timestamps for observation.}
    
    \STATE{$waypoints \gets planningAlgorithm(env)$}
    \STATE{$edges \gets connect(waypoints)$}
    \STATE{$timestamp \gets 0$}
    
    \WHILE{$timestamp \leq observationTimePeriod$}
        \STATE{$timestamp \gets timestamp+1$}
        \STATE{$countDist \gets countDist+collisionCheck(edges) $}
        \STATE{$countTime_{i} \gets countTime_{i} + collisionCheck(edge_i)$ $ \forall i \in $ \{$1,2,....count(edges)$\}}
    \ENDWHILE
    
    \STATE{$D \gets sum(dist(edges))$}
    
    \STATE{$\lambda_{X} \gets countDist/(n*D)$}
    \STATE{$\lambda_{Y_i} \gets (countTime_{i}/n)$ $ \forall i \in $ \{$1,2,....count(edges)$\}}
    \end{algorithmic}
    \label{algo:Poisson}
\end{algorithm}

\subsection{Algorithms}
\begin{figure}
 \centering
 \begin{subfigure}{0.2\textwidth}
 \centering
 \includegraphics[width=0.75\textwidth]{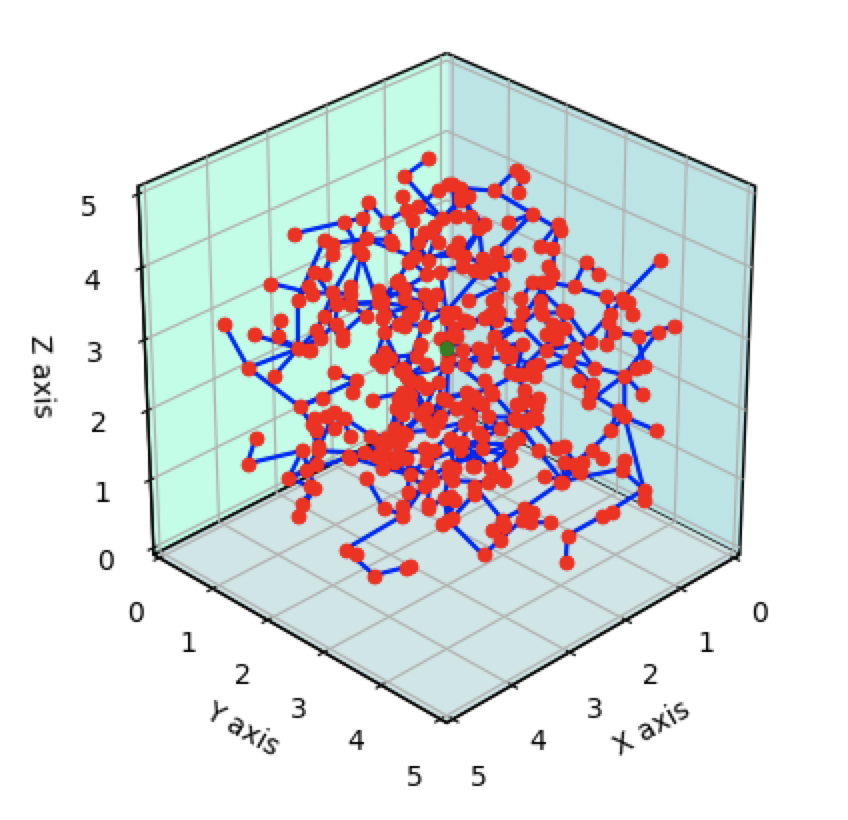}
   \caption{\textbf{RRT* Tree Generation}}
  \label{rrt_star_tree_generation}
 \end{subfigure}%
 \hspace{0.3cm}
 \begin{subfigure}{0.2\textwidth}
 \centering
 \includegraphics[width=0.75\textwidth]{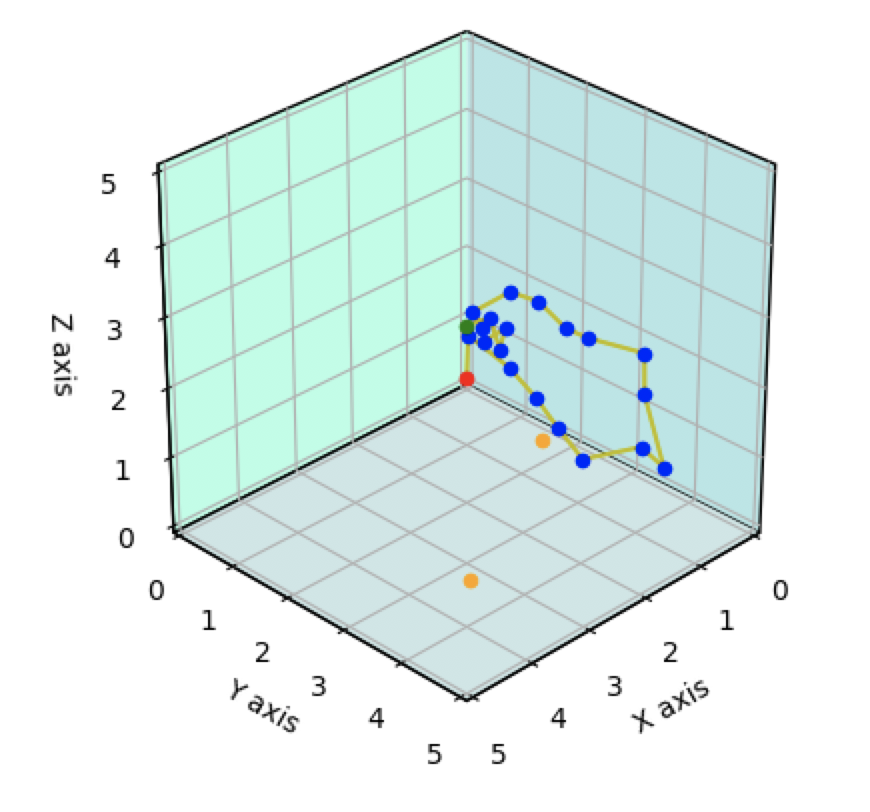}
  \caption{\textbf{Resultant Trajectory Formed}}
  \label{trajectory_formed}
 \end{subfigure}%
 \caption{Trajectory Generation Using RRT*}
 \end{figure}

Algorithm \ref{algo:Poisson} shows an overview of the offline training process. Initially, a trajectory has been generated using an RRT* algorithm as shown in \Cref{rrt_star_tree_generation,trajectory_formed}. This framework can be used with any offline motion planning algorithm without the loss of generality. Once the trajectory has been generated, it then identifies individual waypoints in the C-Space that are part of the trajectory (Step 1 of Algorithm \ref{algo:Poisson}) and connects them using a straight line local planner (Step 2 of Algorithm \ref{algo:Poisson}). It then observes the edges for collision (Steps 4-7 of Algorithm \ref{algo:Poisson}). During the observation period, two types of events are recorded -- one for collision count over the entire trajectory at different timestamps, another for collision count over each edge over a period of time. The algorithm then models them using two Poisson random variables: i.) for the rate of collision per unit distance ($X$) (Step 9 of Algorithm \ref{algo:Poisson}), ii.) for the rate of collision per unit time for each edge connecting two waypoints ($Y$) (Step 10 of Algorithm \ref{algo:Poisson}).

\begin{algorithm}
\caption{Online Planning}
\begin{algorithmic}[1]
    \INPUT{$edges$ determined in \Cref{algo:Poisson}.}
    \INPUT{$\lambda_X$ is the Poisson parameter for the spatial random variable.}
    \INPUT{$\lambda_{Y_i}$ is the Poisson parameter for the temporal random variable for each $edge_i$ in edges.}
    
    \STATE{$threshold \gets 0.2$}
    \STATE{$robotSpeed \gets initialSpeed$}
        
        \FOR{i=1 to count(edges)}
            
            \STATE{$\lambda_{X'} \gets \lambda_X*dist(edge)$}
            \STATE{$traversalTime \gets dist(edge_t)/robotSpeed$}
            \STATE{$\lambda_{Y'} \gets \lambda_Y*traversalTime$}
            \IF{$(P(X'>0) \geq threshold)$}
                \WHILE{$(P(T_{Y'} \leq traversalTime) \geq threshold)$}
                    \STATE{decrease $robotSpeed$}
                    \STATE{$traversalTime \gets dist(edge_t)/robotSpeed$}
                    \STATE{$\lambda_{Y'} \gets \lambda_Y*traversalTime$}
                \ENDWHILE
            \ENDIF
            
            \STATE{traverse the edge at $robotSpeed$}
            \STATE{$robotSpeed \gets initialSpeed$}
        \ENDFOR
        \end{algorithmic}
    \label{algo:Poisson_planning}
\end{algorithm}

Algorithm \ref{algo:Poisson_planning} shows an overview of the online planning process. Initially, it calculates the probability of collision over each edge based on the length of the edge. It also calculates the probability of the arrival time of the next obstacle through that edge being greater than the time taken by the robot to traverse that distance. If both of these probabilities are less than a predefined threshold, the robot is allowed to move to the next waypoint by traversing that edge; otherwise, the speed of the robot is decreased (Step 9 of \Cref{algo:Poisson_planning}). Since the distance between any two consecutive waypoints is fixed, we only manipulate the time it takes for the robot to travel that distance, by manipulating the speed of the robot. This process is repeated (loop in Step 8 of \Cref{algo:Poisson_planning}) until a parameter ($\lambda_Y'$) is found for which the probability of collision is low and it is safe for the robot to move to the next waypoint.
\xspace
\xspace
\begin{table}[!htb]
\begin{minipage}[b]{0.5\textwidth}
\begin{tabular}{|p{1.5cm}|p{1.5cm}|p{1.5cm}|p{1.5cm}|}
	\hline
	\textbf{Moving obstacles} & \textbf{Possible collisions} & \textbf{Collisions avoided} & \textbf{Percentage Accuracy}\\
	\hline
    5 & 18 & 17 & 94.44\\
    \hline
    10 & 28 & 26 & 92.85\\
    \hline
    15 & 41 & 35 & 77.78\\
    \hline
    20 & 60 & 60 & 100\\
    \hline
    25 & 101 & 95 & 95\\
    \hline
    30 & 109 & 103 & 98.09\\
    \hline
    35 & 129 & 122 & 94.57\\
    \hline
    40 & 103 & 97 & 94.18\\
    \hline
    45 & 78 & 71 & 91.02\\
    \hline
    50 & 133 & 124 & 93.23\\
    \hline
\end{tabular}
\caption{\textbf{Predictive Collision Detection Run With Different Numbers of Moving obstacles and 10 Observation timestamps}}
\label{table_results_20_obs}
\end{minipage}
\vspace{2mm}

\begin{minipage}[b]{0.5\textwidth}
\begin{tabular}{|p{1.5cm}|p{1.5cm}|p{1.5cm}|p{1.5cm}|}
	\hline
	\textbf{Moving obstacles} & \textbf{Possible collisions} & \textbf{Collisions avoided} & \textbf{Percentage Accuracy}\\
	\hline
    
    5 & 8 & 7 & 87.5\\
    \hline
    10 & 35 & 32 & 91.43\\
    \hline
    15 & 60 & 56 & 93.33\\
    \hline
    20 & 61 & 55 & 90.16\\
    \hline
    25 & 84 & 77 & 91.67\\
    \hline
    30 & 93 & 86 & 92.47\\
    \hline
    35 & 179 & 162 & 90.50\\
    \hline
    40 & 146 & 135 & 92.47\\
    \hline
    45 & 144 & 137 & 95.14\\
    \hline
    50 & 164 & 156 & 95.12\\
    \hline
\end{tabular}
\caption{\textbf{Predictive Collision Detection Run With Different Numbers of Moving obstacles and 20 Observation timestamps}}
\label{table_results_10_obs}
\end{minipage}
\end{table}
\xspace
 \begin{figure*}
 \centering
 \begin{subfigure}{0.2\textwidth}
 \centering
 \includegraphics[width=\textwidth]{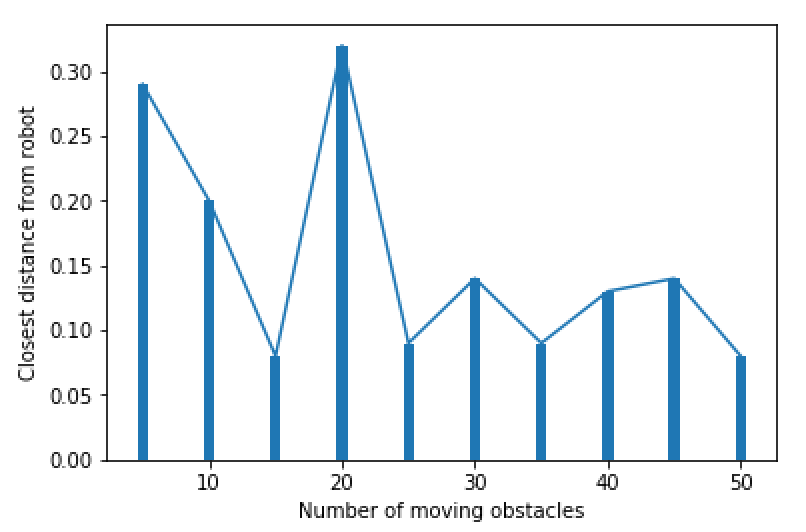}
   \caption{\textbf{Closest distance of obstacles from robot with 10 observation timestamps}}
  \label{closest_dist_10_ts}
 \end{subfigure}%
 \hspace{0.3cm}
 \begin{subfigure}{0.2\textwidth}
 \centering
 \includegraphics[width=\textwidth]{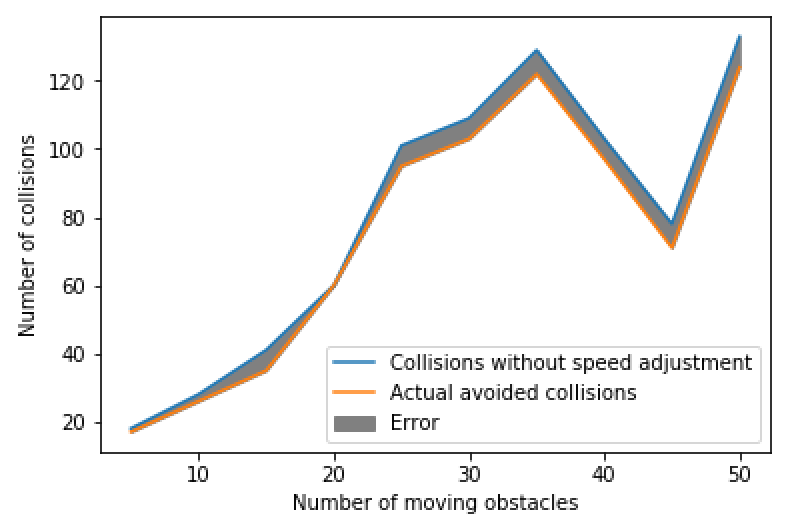}
  \caption{\textbf{Collisions with and without speed adjustment with 10 observation timestamps}} 
  \label{num_collision_10_ts}
 \end{subfigure}%
 \hspace{0.3cm}
  \begin{subfigure}{0.2\textwidth}
 \centering
 \includegraphics[width=\textwidth]{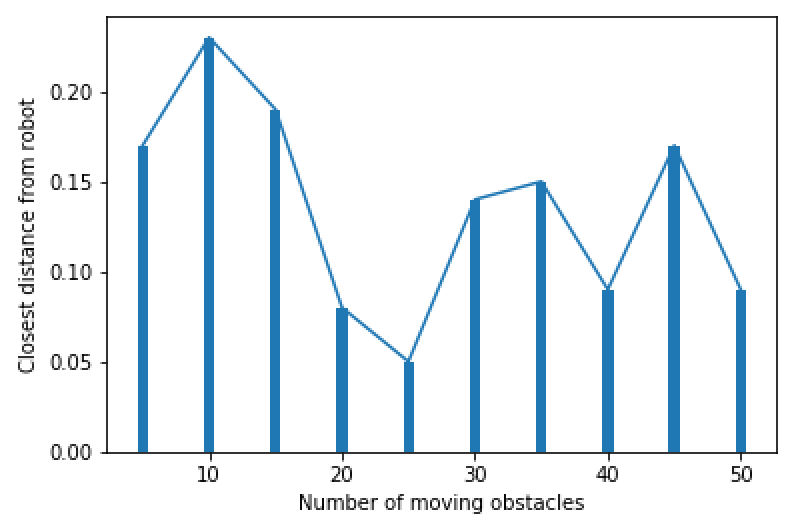}
   \caption{\textbf{Closest distance of obstacles from robot with 20 observation timestamps}}
  \label{closest_dist_20_ts}
 \end{subfigure}%
 \hspace{0.3cm}
 \begin{subfigure}{0.2\textwidth}
 \centering
 \includegraphics[width=\textwidth]{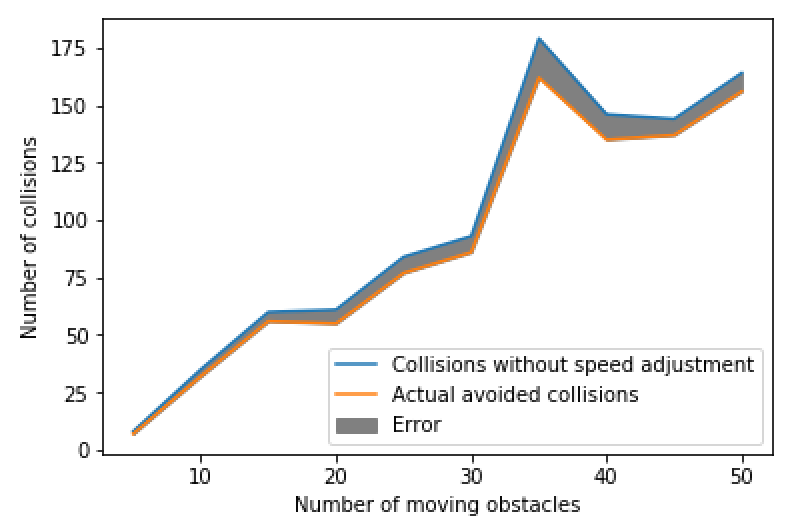}
  \caption{\textbf{Collisions with and without speed adjustment with 20 observation timestamps}} 
  \label{num_collision_20_ts}
 \end{subfigure}%
 \caption{Simulation experiments with 10 and 20 observations}
 \label{simulation_graphs}
 \end{figure*}
\xspace
\begin{figure*}
\centering
\begin{minipage}[b]{\textwidth}
\centering
\includegraphics[width=\textwidth]{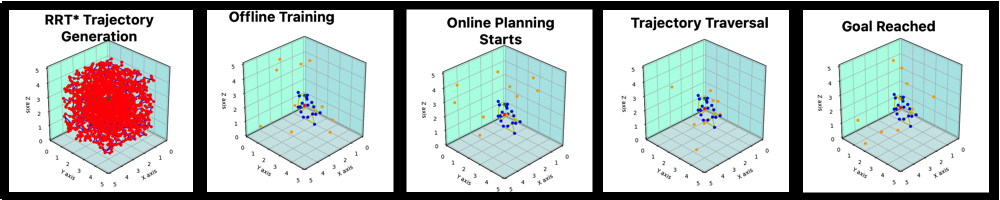}
  \caption{\textbf{Simulation experiment on an environment with 10 random moving obstacles }(marked as floating dots without any edges)}
\label{film_strip_simulation}
\end{minipage}
\end{figure*}
\xspace
\subsection{Complexity analysis}
The worst case time complexity for the \Cref{algo:Poisson} is $O(m*n)$ where m is the number of edges on the trajectory and n is the number of observations. The worst case time complexity for the \Cref{algo:Poisson_planning} is $O(m*o)$ where o is the number of times we have to decrease the speed of the robot. Given a large value of m, we can simplify the time complexity to be $O(m^2)$.


\section{Experiments}
In order to demonstrate the proficiency of the approach, we compared it with other existing algorithms that solve the same problem. 
The evaluation has been done in stages. A proof of concept involving simulation experiments has been performed. The simulation experiments were done using a python simulation environment built using pygame \cite{pygame}. The success of this proof of concept leads to real-world testing.

\subsection{Experimental Setup}
In order to simulate a 3D environment, 3D plotting libraries were used in python. The moving obstacles were generated at random coordinates to make their trajectories unpredictable. The real-world testing used crazyflie 2.1 drones from bitcraze \cite{bitcraze} in a controlled setting utilizing controlled moving obstacles. The moving obstacles were other crazyflie drones that were following a fixed trajectory. All the drones in the environment were communicating with the bitcraze Loco Positioning System (LPS).

\begin{figure}
\label{Real_world_experimental_setup}
\centering
\begin{minipage}[b]{0.3\textwidth}
\centering
\includegraphics[width=\textwidth]{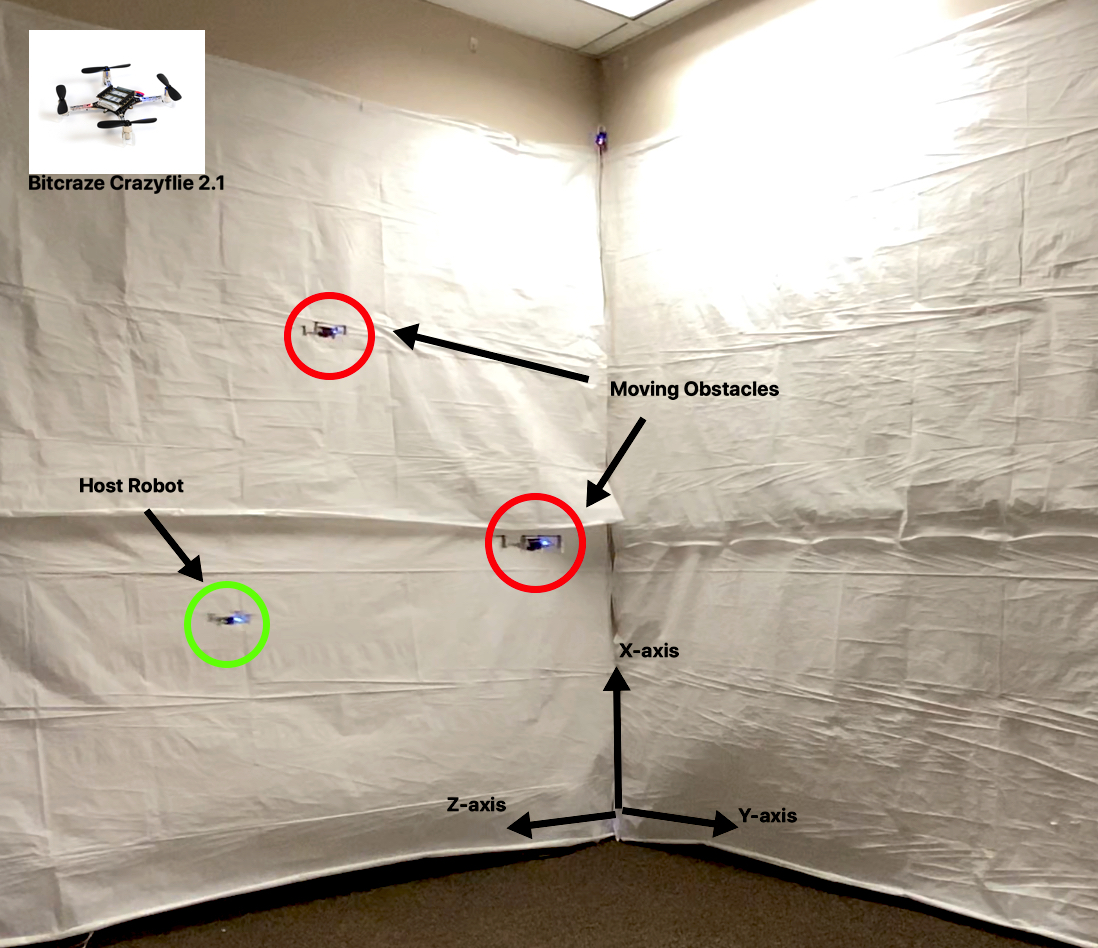}
  \caption{\textbf{Real world experimental setup}}
\end{minipage}
\end{figure}

A typical real-world controlled experiment would involve running a Ubuntu 18.04 operating system on a Dell Optiplex computer with a Crazy-radio antenna (also from bitcraze) receiving signals from 8 LPS-anchor tags at each corner of the 3D environment to estimate the position of the drones in the environment. The environmental setup is shown in \Cref{Real_world_experimental_setup}.

\subsection{Experimental Results}

  \begin{figure*}
\centering
\begin{minipage}[b]{\textwidth}
\centering
\includegraphics[width=\textwidth]{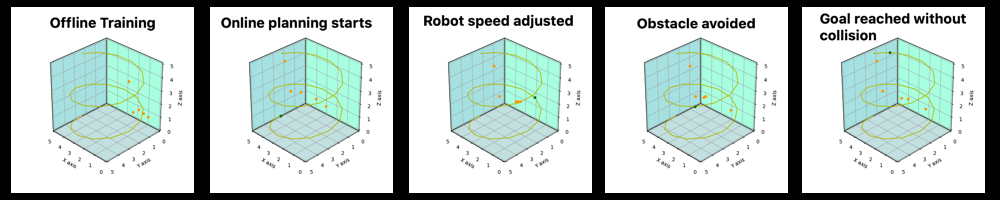}
  \caption{\textbf{Simulated representation of real-world experiment on an environment with 5 moving obstacles }(marked as floating dots without any edges) \textbf{ with predefined trajectories and different speeds} }
\label{film_strip_real}
\end{minipage}
\end{figure*}

\subsubsection{\textbf{Simulation experiments}}
We run 10 experiments each with increments of 5 obstacles. We also recorded the results from 10 and 20 observations timestamps for the training algorithm. \Cref{table_results_10_obs} and \ref{table_results_20_obs} lists the results from the experiments run on different environments with varying numbers of moving obstacles with 10 and 20 observations timestamps. As we can see, the number of possible collisions without any speed adjustment (control experiment) increases as the number of moving obstacles increases, but the number of avoided obstacles also improved. This has happened because, with more obstacles in the environment, the Poisson processes were trained better than with a lower number of obstacles. \Cref{closest_dist_10_ts} and \ref{closest_dist_20_ts} show the plot of closest distances of any obstacle to the robot with respect to the number of moving obstacles. \Cref{num_collision_10_ts} and \ref{num_collision_20_ts} plot the number of collisions with and without the collision avoidance algorithm with 10 and 20 observation timestamps, respectively. It also shows the margin of error for each experiment.
The robot had no knowledge of the total number of collisions; it has been determined for recording the accuracy of the algorithm only after its execution has completed. \Cref{film_strip_simulation} shows the step by step execution of our algorithm on a simulated environment with 20 moving obstacles generated randomly.

The average closest distance when the robot's maximum speed is $1 m/s$ was $0.156$ when only 10 observation timestamps were used to train the algorithm with 5 to 50 moving obstacles. When compared to the closed-loop RRT algorithm presented in \cite{lin2017sampling} that uses up to 2 moving obstacles, the average closest distances were $2.35$ for speed of $1 m/s$. Our environment is of size $5*5*5$ cubic meters, thus, we scale the average distance between the robot and the moving obstacle, to make a fair comparison. The environment size used in \cite{lin2017sampling} was approximately $15*15*15$ cubic meters. We scale the average closest distance between the robot and the moving obstacle from our algorithm by a factor of $27$. In a scaled environment of similar size as \cite{lin2017sampling}, we would get an average closest distance of $4.212$ meters for speeds of $1 m/s$. Thus our algorithm performs better with the larger closest distance.

\subsubsection{\textbf{Real-world experiments}}

In the real-world experiments, we consider a 0.3 meters proximity between host drone (drone running the predictive collision algorithm) and moving obstacle as a collision. This distance is optimum because at this distance the obstacle drone can be avoided and the host drone can continue through its trajectory even if it has failed to predict a collision. A test run is considered to be successful if it has less than 20\% of unpredicted collisions. With up to two moving obstacles, we found that the trajectory of the host drone was successfully traversed without colliding with the moving obstacles. Refer to the attached video for details. A set of control experiments were conducted to test whether the host drone can traverse without the use of our algorithm. In all the cases, the host and the obstacle drones collided and the experiment failed, without the use of predictive collision avoidance. This necessitates the use of our algorithm for dynamic obstacle avoidance. \Cref{film_strip_real} shows the step by step execution of our algorithm on a simulated representation of a real-world environment with 5 moving obstacles, where, both the robot and the obstacles had a predefined trajectory. The robot (in blue) follows an uprising spiral trajectory while the obstacles (in orange) follow a straight line parallel to the X-axis with different speeds, and the robot slows down due to our algorithm.
\xspace
\xspace
\section{Conclusion}
We have developed a dynamic collision detection and avoidance framework that is compatible with most of the motion planning strategies. We have tested with a popular algorithm (RRT*) and achieved great results in terms of accurate prediction of collisions. We have shown that by using our framework, there is no necessity for dynamic-replanning for the trajectory. We have used a static trajectory, as shown in the Algorithm \ref{algo:Poisson}. As future work, we would like to study if Poisson processes trained offline can be used to generate static trajectories in dynamic environments by calculating collision probabilities during local planning.
\xspace
\bibliography{predict}
\bibliographystyle{plain}
\end{document}